\documentclass{article}

\usepackage{amssymb}
\usepackage{fancybox}  
\usepackage[breakable]{tcolorbox}
\usepackage{graphicx}
\usepackage{tcolorbox}
\usepackage{enumitem}
\usepackage[sort,numbers]{natbib}
\usepackage{fontawesome5} 
\usepackage{xcolor}
\usepackage{caption}
\usepackage{arxiv}
\usepackage{hyperref}
\usepackage[hang,flushmargin]{footmisc}
\usepackage{enumitem}
\setlist[itemize]{parsep=1pt}
\title{PCS Workflow for Veridical Data Science\\
in the Age of AI}

\author{Zachary T. Rewolinski\\
        Department of Statistics\\
        UC Berkeley\\
        \texttt{zachrewolinski@berkeley.edu}\\
        \And
        Bin Yu\\
        Department of Statistics and EECS\\
        UC Berkeley\\
        \texttt{binyu@berkeley.edu}}

\begin{document}

\maketitle
\begin{abstract}

Data science is a pillar of artificial intelligence (AI), which is transforming nearly every domain of human activity, from the social and physical sciences to engineering and medicine.
While data-driven findings in AI offer unprecedented power to extract insights and guide decision-making, many are difficult or impossible to replicate.
A key reason for this challenge is the uncertainty introduced by the many choices made throughout the data science life cycle (DSLC).
Traditional statistical frameworks often fail to account for this uncertainty.
The Predictability-Computability-Stability (PCS) framework for veridical (truthful) data science offers a principled approach to addressing this challenge throughout the DSLC.
This paper presents an updated and streamlined PCS workflow, tailored for practitioners and enhanced with guided use of generative AI.
We include a running example to display the PCS framework in action, and conduct a related case study which showcases the uncertainty in downstream predictions caused by judgment calls in the data cleaning stage.

\end{abstract}
\section{Introduction}

Data analysis is a human activity as ancient as agriculture and trade.
Today's data analysis takes many forms across different fields.
It varies in terms of data size, data quality, available computing resources, domain knowledge, and the level of human engagement.
These differences are shaped by the problems being addressed in areas such as science, medicine, engineering, and more.
Over the last century, human analysts have delegated more and more responsibilities to software.
Whereas linear models were once fit by written calculations, now tools like \texttt{python} and \texttt{R} automate these computations.
With the advent of generative AI (GenAI) systems such as ChatGPT, Claude, and Gemini, human-machine collaboration in data science is expected to enter a new era.

Data analysis as practiced today is (implicitly or explicitly) carried out through a process we call the data science life cycle (DSLC) \cite{box1976science}.
The DSLC is a structured investigation process rooted in the context of a specific domain problem.
Each step in the DSLC requires human ``judgment calls" pertaining to domain knowledge and algorithmic approaches.
These decisions can vary between individuals or teams, as they often involve subjective interpretations, assumptions, or preferences.
As a result, different but reasonable choices can lead to different analytical paths and outcomes, introducing uncertainty into the process \cite{uncertainty2023yu, gould2025same, breznau2022data}.
Traditional statistical modeling frameworks do not consider the variability stemming from judgment calls.
Under-accounting these sources of uncertainty contributes in a significant way to irreproducibility of results and conclusions \cite{gould2025same, breznau2022data}.

The Predictability-Computability-Stability (PCS) framework for veridical (truthful) data science has been developed by the second author and her collaborators throughout the last decade \cite{stability2013yu, vds2020yu, uncertainty2023yu, veridical2024yu}.
PCS was motivated to meet the challenges set forth by the replication crisis in the early 2010s.
In particular, \citet{prinz2011believe} and \citet{begley2012replication} concluded that only 11-25\% of major discoveries in pre-clinical oncology research at biotech companies Amgen and Bayer could be replicated.
After the PCS framework was introduced in 2020 by \citet{vds2020yu}, it so happened that groups of scientists brought renewed attention to this replication issue.
Recent studies from social science and ecology showed that different teams of qualified analysts' reasonable judgment calls could lead to different data conclusions \cite{breznau2022data, gould2025same}.
This lends compelling evidence for the need of frameworks like PCS, which systematically address and quantify the impact of such judgment calls by explicitly examining the uncertainties they introduce.
In a case study included in Section \ref{sec:casestudy}, students in a graduate applied statistics and machine learning course (STAT 214) at UC Berkeley to clean a medical dataset based on guidance from a clinician.
Students made widely varying judgment calls when handling missing and inconsistent data.
As a result, the predictive performance of models on each cleaned dataset showed uncertainty comparable to that induced by bootstrapping each student's cleaned dataset individually.

PCS-guided approaches have yielded successes in producing trustworthy insights and knowledge.
\citet{dwivedi2020stable} applied PCS principles in medicine to find stable and interpretable subgroups in randomized controlled trials.
In genomics, \citet{wang2024epistasis} developed a low-signal method to find experimentally validated causal genetic drivers for a heart disease called hypertrophic cardiomyopathy (HCM).
Furthermore, \citet{tang2025simplified} developed a PCS ranking procedure to select predictive genes for cost-effective cancer diagnosis.
We note that while these successes are predominantly in the biomedical sciences due to the Yu Group's close collaborations with domain experts in medicine, the PCS framework is equally applicable across domains.

In this paper, we provide an  updated and streamlined PCS workflow for practitioners and researchers making data-driven conclusions.
This approach builds on the work of \citet{vds2020yu} and \citet{veridical2024yu} while also considering the emerging contributions and implications of GenAI.
The remainder of this work is structured as follows.
Section \ref{sec:framework} describes the PCS framework and its core principles.
Sections \ref{sec:formulation}-\ref{sec:communication} provide step-by-step instructions for the DSLC according to these PCS principles, including tips and warnings for incorporating the use of GenAI.
Section \ref{sec:casestudy} presents the case study described above, which uses an example from the authors' STAT 214 course to examine uncertainty resulting from judgment calls in data cleaning.
We conclude with a brief discussion on veridical data science and opportunities for future work.
\section{The PCS Framework for Veridical Data Science}\label{sec:framework}

\begin{figure}
    \centering
    \includegraphics[width=\linewidth]{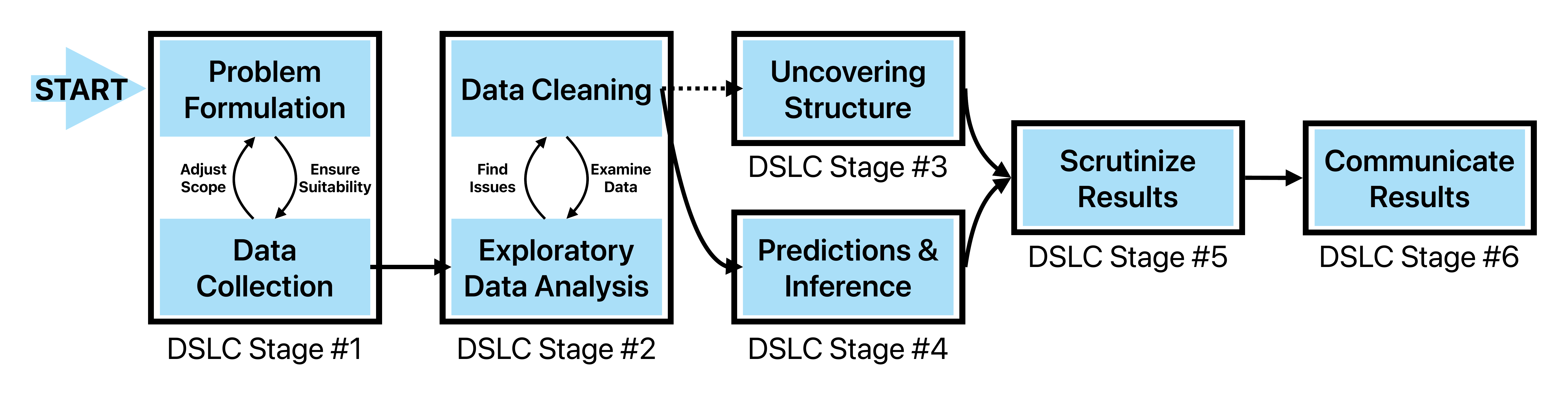}
    \caption{The six stages of the data science life cycle.}
    \label{fig:workflow}
\end{figure}

PCS is an evolving conceptual, philosophical, and practical research program for conducting and documenting veridical data science and AI.
It rests on three core principles of data science: \textbf{P}redictability, \textbf{C}omputability, and \textbf{S}tability \cite{vds2020yu}.
These principles represent an integration of best practices and ideas from the two cultures of machine learning and statistics \cite{breiman2001cultures}.
Furthermore, the PCS framework includes rigorously documenting the justification for decisions made throughout the DSLC using code, visualizations, and writing.
An example of PCS-adherent documentation can be found in \citet{tangvdocs}.
PCS for veridical data science builds ``one culture" with unified language to ensure trustworthiness and safety of data science and AI.

\textbf{P} and \textbf{C} are main tenets of machine learning (ML).
\textbf{P} is used to ensure that the DSLC adequately captures reality and accurately reflects the goal of the domain problem.
We thus refer to the use of \textbf{P} as a \textit{reality check} \footnote{\fontsize{7.25pt}{10pt}\selectfont \textbf{P}redictability is used rather than \textbf{R}eality since early-stage development of PCS used \textbf{P} for supervised learning.}.
In the supervised learning setting, predictions are a natural way to perform such a reality check \cite{vds2020yu}.
\citet{veridical2024yu} recently extended the reality check to unsupervised learning and the rest of the DSLC by using qualitative domain knowledge and/or cross-validating discovered structure in held-out data.
\textbf{C} goes beyond computational considerations such as speed and memory to include data-inspired simulations, as outlined in the PCS-guided simulation protocols MERITS and simChef \cite{elliott2024designing, Duncan2024simChef}.

\textbf{S} ensures that data-driven results remain stable under reasonable perturbations to the DSLC, where ``reasonable" is defined within the context of the domain problem and described in documentation.
This represents a substantial contribution of PCS toward establishing trust in data conclusions.
As shown in Figure \ref{fig:workflow}, the DSLC is an iterative process consisting of six stages.
These stages begin by transforming a domain problem into a data science problem and end with communicating results effectively to a target audience.
Within each stage, \textbf{S} involves using context-appropriate metrics to measure the impacts of reasonable perturbations.
Such \textit{stability checks} should be performed as frequently as resources allow.
Each stability check should subsequently be thoroughly documented.
Examples of suitable perturbations might include using UK BioBank data instead of Japan BioBank data or using a different but reasonable data cleaning procedure.
In the modeling phase, perturbations may involve using random forests instead of gradient boosting for prediction, or using different measures of feature importance to explain the model.
One special form of stability analysis is using PCS prediction perturbation intervals to report uncertainty quantification rather than relying on a single point estimate.
These intervals were formally introduced for regression in \href{https://vdsbook.com/13-final_ml}{Chapter 13} of \citet{veridical2024yu} and for multi-class classification in recent work by \citet{agarwal2025pcsuq}.
In both exploratory data analysis and presenting results, stability checks may entail using different types of visualizations.
If sources of instability are found, \textbf{S} furthermore involves improving the stability of the DSLC \footnote{\fontsize{7.25pt}{10pt}\selectfont Stability improvements using problem-specific information include \citet{basu2018iterative}, \citet{wang2024epistasis}, \& \citet{tang2025simplified}.}.
Thus, \textbf{S} is a significant expansion of the sample-to-sample perturbations considered in statistics under a probabilistic generative model.

In hindsight, PCS is common sense.
It simply says ``let us aggregate reality-checked data analysis approaches".
This is reflective of how in real life, we may say ``let us combine the opinions of good doctors" when faced with a serious illness.

\subsection{PCS Use of Generative AI in Data Science}

Data scientists today are frequently turning to GenAI for support throughout each stage of the DSLC.
Although GenAI systems such as large language models (LLMs) can be powerful, they also have well-documented downsides.
A major concern is the tendency of LLMs to hallucinate, a phenomenon in which plausible yet false information is confidently presented as fact \cite{huang2025hallucination}.
Thus, trustworthy use of GenAI demands adherence to PCS principles in its deployment.
For example, LLMs such as Gemini or Llama may be a useful way to quickly gain a high-level understanding of unfamiliar subjects relevant to the data science task.
A PCS-aligned method for mitigating hallucinations is cross-referencing different GenAI systems by giving similar prompts to multiple chatbots \cite{chiang2024chatbot}.
Another suggested stability check is repeating the prompt with small perturbations (while keeping the overall message the same) in order to~determine~if the answer given by an LLM is consistent.
Lastly, GenAI output should always be verified with trusted external~sources.

To mitigate potential problems and ensure truthful data science, we provide recommendations for using GenAI tools at the end of the following sections.
To the best of our knowledge, this is the first formal guidance on incorporating GenAI use into the DSLC.
We note that with the rapid advancement of GenAI's capabilities, the precise recommendations given might not hold in a few years' time.
However, the general PCS spirit of these recommendations will continue to hold.

\section{DSLC Stage \#1: Problem Formulation and Data Collection}
\label{sec:formulation}

The DSLC begins by working alongside domain experts to define a problem of interest which would benefit from data-driven insights and conclusions.
While this may seem trivial, the initial domain problem can often be too vague to be properly answered.
To better form a complete question, the data scientist must grasp the underlying goals of the domain experts.
This may require the data scientist to gain an understanding of the domain of interest.

After these goals are clear, the next step is to find available data relevant to the task.
If an insufficient amount of data exists, either additional data must be gathered or the question must be adjusted.
We encourage the involvement of data scientists in the data collection process when possible.
This will help ensure the data is properly formatted, thoroughly documented, and able to provide strong evidence towards a conclusion.
When existing data is used, we likewise encourage data scientists to understand the steps taken in data collection and how they may impact any downstream tasks.
It is essential that existing data has sufficient documentation to understand its intricacies.
Further~readings on problem formulation and data collection can be found in \href{https://vdsbook.com/02-dslc#dslc-stage-1-problem-formulation-and-data-collection}{Chapter 2.2} of \citet{veridical2024yu}.

\subsection{Running Example: Clinically Important Traumatic Brain Injuries}
\label{subsec:citbi}

Here we introduce a real-world example that we will return to throughout this work.
Traumatic brain injury (TBI) from blunt trauma is caused by an external force which moves the brain within the skull.
TBI is a major cause of death in children \cite{langlois2006tbi, faul2010tbi}.
Thus, children with clinically-important traumatic brain injury (ciTBI) requiring intervention such as neurosurgery need to be quickly identified.
Computed tomography (CT) scans are the standard of care for diagnosing TBI \cite{doezema1991ct}.
However, CT scans expose the patient to ionizing radiation.
A recent study from \citet{smith-bindman2025cancer} found that radiation-induced cancer is estimated to result from one in every 313 head CT scans given to children.
Furthermore, head CT scans contributed the largest number of cancers in children (53\%) compared to other CT categories.
Since only about 5\% of CT scans find injuries, there are many CT scans causing needless risk to patients \cite{kuppermann2009tbi}.
Thus, identifying children with near-zero risk of ciTBI can improve patient outcomes by reducing unnecessary CT scans.
This was the main goal of \citet{kuppermann2009tbi}, a nationwide collaboration among physicians and researchers across various hospitals and universities in the Pediatric Emergency Care Applied Research Network (PECARN).

To arrive at a data-driven conclusion, the PECARN team decided to collect new clinical data regarding the outcomes of children with mild head trauma.
A detailed collection plan was created, including important considerations such as patient follow-up and quality assurance across a network of 25 emergency departments.
A \href{https://pecarn.org/studyDatasets/documents/TBIDataForms1-4-6-7.pdf}{standardized data collection form} would be filled in by a physician, nurse practitioner, or physician assistant while visiting the patient in the emergency department.
Detailed data documentation was included to assist researchers in the analysis of the collected data.
The resulting data and documentation was made publicly available online.
We consider this to be a strong example of thorough data collection, with many best practices being followed.

\begin{center}
    \begin{tcolorbox}[colback=yellow!20, colframe=black, width=0.9\linewidth]
    \centering\faLightbulb[regular]\hspace{0.25cm}\textbf{Generative AI Tips \& Warnings}\hspace{0.25cm}\faExclamationTriangle
        \begin{itemize}[leftmargin=0pt]
            \item Give identical prompts to multiple different GenAI systems (Gemini, Claude, ChatGPT, etc.) to ensure stability.
            \item Change the wording of your prompt, while keeping the main message the same. Does it change the response in a meaningful way?
            \item Check reliable external sources support the LLM's output prior to incorporating your domain knowledge into the project or task.
        \end{itemize}
    \end{tcolorbox}
\end{center}
\begin{center}
    \begin{tcolorbox}[width=0.9\linewidth]
    \centering\faClipboardList\hspace{0.25cm}\textbf{Key Takeaways \& Action Items}\hspace{0.25cm}\faClipboardCheck
        \begin{itemize}[leftmargin=0pt]
            \item Ensure the data-driven task is aligned with the goals of domain experts.
            \item Verify that existing or collected data is sufficient to answer the given question.
            \begin{itemize}
                \item If not, problem formulation will have to be revisited.
            \end{itemize}
            \item Create thorough data documentation throughout the data collection process, or ensure your previously existing data has sufficient documentation.
        \end{itemize}
    \end{tcolorbox}
\end{center}

\section{DSLC Stage \#2: Data Cleaning and Exploratory Data Analysis}
\label{sec:cleaning}

Data cleaning is perhaps the most underemphasized aspect of the DSLC.
Nearly all real-world data requires preparation prior to its use in aiding conclusions.
However, important data cleaning decisions are often swept under the rug without regard to their downstream impacts.

The overarching goal of data cleaning is to obtain the version of your data that will best help reach truthful conclusions.
Achieving this goal requires familiarity with both the domain problem and the intricacies of the data.
Understanding the collection process and data documentation are necessary for trustworthy data cleaning, but not sufficient for full comprehension of the data's complexities.
Thus, we suggest an approach that incorporates exploratory data analysis (EDA) as a tool for thorough data cleaning.
Instead of treating data cleaning like a linear task, it should be an iterative process that evolves as insights are discovered through EDA.
For example, by plotting a histogram to view the distribution of a covariate, the data scientist may notice frequent missingness, suggesting imputation may be helpful for downstream modeling.
Performing a stability check by plotting a boxplot instead may reveal outliers with infeasible values, uncovering possible mistakes in data entry.
With each new discovery, we apply the appropriate data cleaning adjustments and return to EDA, continuing the cycle.
For details such as suggested explorations and cleaning steps, we refer the reader to Chapters \href{https://vdsbook.com/04-data_cleaning}{4} and \href{https://vdsbook.com/05-data_viz}{5} of \citet{veridical2024yu}.

Each judgment call in this process will result in a different cleaned dataset, potentially causing differences in results and conclusions.
We recommend producing multiple versions of cleaned data by changing some of the key judgment calls.
This will allow for a downstream analysis of stability to different data cleaning processes.
If the results are not stable with respect to data cleaning perturbations, the judgment calls will have to be carefully determined and documented.

Once EDA for data cleaning is complete, a polished explanatory version can be produced to communicate interesting patterns and trends to external audiences.
However, caution is needed to avoid drawing conclusions from patterns that may not hold in new or unseen data.
Some fields address this by requiring pre-registration of studies, which can also limit the opportunity for researchers to uncover unanticipated insights.
The PCS approach offers a balanced alternative in which open exploration of the data is encouraged and strong evidence is required before drawing conclusions.
This evidence should include demonstrations of stability across different data cleaning and preprocessing choices.
It should also show stability across reasonable perturbations, such as bootstrapped samples of the data.

\textbf{Example.}
Let us again consider the PECARN ciTBI data.
While \citet{kuppermann2009tbi} approached data collection thoughtfully, there are still difficulties present in the dataset.
Due to the high-stakes setting of the emergency department, many patients have missing or inconsistent entries, which are likely attributable to human error.
By referencing both the data documentation and the data entry form, we can better understand the missingness present in the data.
The documentation notes that many covariates have ``not applicable" encoded as ``92".
This serves as a valuable differentiator between unintentional and intentional missingness.
For example, the dataset includes an indicator for whether trauma above the clavicles is observed.
In addition, six indicators denoting the location (face, neck, etc.) of such trauma are also present.
It is clear from the data entry form that if a patient does not have trauma above their clavicles, all six associated indicators should intentionally have the missing value encoding of ``92".
However, some measurements may be left blank, indicating unintentional missingness potentially attributable to human error.
We address uncertainty stemming from data cleaning judgment calls on the ciTBI dataset more concretely in Section \ref{sec:casestudy}.

\begin{center}
    \begin{tcolorbox}[colback=yellow!20, colframe=black, width=0.9\linewidth]
    \centering\faLightbulb[regular]\hspace{0.25cm}\textbf{Generative AI Tips \& Warnings}\hspace{0.25cm}\faExclamationTriangle
        \begin{itemize}[leftmargin=0pt]
            \item LLMs can assist in efficiently creating multiple copies of clean data by providing starter code. It is important that the analyst verifies the generated code.
            \item Coding assistants such as GitHub Copilot or Cursor may be helpful for documenting and commenting data cleaning code.
        \end{itemize}
    \end{tcolorbox}
\end{center}
\begin{center}
    \begin{tcolorbox}[width=0.9\linewidth]
    \centering\faClipboardList\hspace{0.25cm}\textbf{Key Takeaways \& Action Items}\hspace{0.25cm}\faClipboardCheck
        \begin{itemize}[leftmargin=0pt]
            \item Decisions made during the cleaning process should be thoroughly documented.
            \item Data cleaning should be an ongoing process that evolves as insights are uncovered through exploratory data analysis (EDA).
            \begin{itemize}
                \item Use EDA to identify patterns, outliers, and inconsistencies that may require further refinement of the dataset. Be cautious of over-interpreting patterns that might not generalize.
            \end{itemize}
            \item Keep multiple versions of cleaned data for use in downstream stability checks.
        \end{itemize}
    \end{tcolorbox}
\end{center}

\section{DSLC Stage \#3: Uncovering Data Structure with Unsupervised Learning}
\label{sec:structure}

In the DSLC, exploring the underlying structure of the data is an optional yet valuable step.
Identifying interesting patterns or structures within the data can uncover insights that may guide the subsequent analysis.
In some cases, discovering new structures could provide important takeaways or results that influence the overall direction of the project.

Common unsupervised learning approaches include dimensionality reduction and clustering.
These techniques present a challenge, since the lack of outcome labels prevents reality checks which leverage predictions.
Dimensionality reduction methods such as principal component analysis (PCA) \cite{hotelling1933analysis} help simplify the task at hand by finding a handful of features that explain the majority of the variance in the data.
Stability checks of the selected features can be performed by ensuring their robustness under different data cleaning and preprocessing decisions.
If labels are available, performing a principal component regression can serve as a reality check, ensuring that the features are consistent and informative for the given problem.
Similarly, subgroups found by clustering methods can be assessed for consistency across multiple algorithms, including hierarchical clustering, k-means clustering, and t-SNE, to name a few \cite{ward1963cluster, lloyd1982least, macqueen1967some, vandermaaten2008tsne}.
Furthermore, hyperparameters such as the number of clusters can be chosen by their ability to produce stable groupings.
For example, a common approach is to evaluate the Jaccard similarity \cite{jaccard1901index} of clusters obtained from different random subsamples \cite{BenHur2002stability}.
Consistent subgroup structure provides a valuable reality check, helping ensure associated findings are robust and reliable.
PCA and clustering are described in greater detail in Chapters \href{https://vdsbook.com/06-pca}{6} and \href{https://vdsbook.com/07-cluster}{7} of \citet{veridical2024yu}.

\textbf{Example.}
The PECARN team's goal was to identify and characterize the group of patients with near-zero risk of ciTBI.
Uncovering any natural subgroup structure amongst patients may aid in detecting such a group.
In the case where no stable clusters are discovered, the information gained by the resulting exploration still helps build familiarity with our data, much like EDA.
However, we note that \citet{kuppermann2009tbi} did not use unsupervised learning, opting instead to skip this stage.
While this choice is not discussed in their work, a possible explanation is that unsupervised learning techniques are difficult to apply to cases with missingness or categorical features, both of which are heavily present in the ciTBI data.

\begin{center}
    \begin{tcolorbox}[colback=yellow!20, colframe=black, width=0.9\linewidth]
    \centering\faLightbulb[regular]\hspace{0.25cm}\textbf{Generative AI Tips \& Warnings}\hspace{0.25cm}\faExclamationTriangle
        \begin{itemize}[leftmargin=0pt]
            \item Do not solely rely on GenAI to generate pipelines for complex techniques such as PCA or clustering. Always ensure that GenAI-written code is correct.
            \begin{itemize}
                \item Understanding how these methods work helps you verify that the coding assistant is using them properly.
            \end{itemize}
        \end{itemize}
    \end{tcolorbox}
\end{center}
\begin{center}
    \begin{tcolorbox}[width=0.9\linewidth]
    \centering\faClipboardList\hspace{0.25cm}\textbf{Key Takeaways \& Action Items}\hspace{0.25cm}\faClipboardCheck
        \begin{itemize}[leftmargin=0pt]
            \item Determine if the domain problem would benefit from a certain type of underlying data structure.
            \item Incorporate stability and predictability checks into unsupervised learning to ensure validity of findings.
        \end{itemize}
    \end{tcolorbox}
\end{center}

\section{DSLC Stage \#4: Predictive Modeling }
\label{sec:predictions}

In the context of statistical modeling, it’s important to draw a distinction between two core concepts: \textbf{predictions} and \textbf{inference}.
Predictions refer to the task of predicting responses of unobserved observations based on a trained model, while inference involves the estimation of unknown population parameters.
The key difference is that (parameter) inference typically relies on distributional assumptions that may not checkable.
The development of PCS thus far has focused on predictions rather than inference.
Within the domain of predictions, the PCS framework can be divided into two distinct categories: model training and generating predicted responses.
The first category focuses on ensuring the model itself is aligned with reality and stable throughout its development, while the second revolves around using the trained model to attain reliable predictions.
For more details on specific types of predictive models, we refer the readers to Chapters \href{https://vdsbook.com/08-prediction_intro}{8}-\href{https://vdsbook.com/12-rf}{12} of \citet{veridical2024yu}.

\subsection{PCS Training of Models}

Our goal is to ensure that models are PCS-aligned throughout their development.
The PCS framework emphasizes frequent reality and stability checks for judgment calls such as model selection or hyperparameter tuning.
A key component in this process is pre-selecting a stability evaluation metric.
Accuracy metrics such as $R^2$, F1 score, mean squared error, and more are chosen based on the problem at hand.
For example, in the case of the PECARN ciTBI task, the researchers are primarily interested in maintaining a low false negative rate.
This makes sense for the applied problem, since the ``worst" outcome for the emergency department is the case where a ciTBI diagnosis is missed and the patient dies as a result.

We argue that similar to selecting a measure for accuracy, data scientists should choose a stability metric beforehand which aligns with the specific prediction task.
Stability metrics allow practitioners to assess how consistent the model's performance is when subjected to different data perturbations.
Data perturbations are a powerful tool for ensuring stability in model selection.
By applying perturbations to the input data (such as adding noise or making small adjustments), one can test how sensitive the model is to changes in the data.
For more information on how to perform perturbations to evaluate model stability, we refer the reader to \citet{vds2020yu}.

While the predominant approach in modern data science is to choose the one ``optimal" model based on performance on a held-out test set, we advocate for the use of multiple models---whether this means different algorithms altogether, the same predictive algorithm with different hyperparameter selections, or using different versions of cleaned data to train the model.
By considering a variety of predictive models, we can better understand the uncertainty that can appear in our predictions.
These different models may benefit from having different amounts of complexity.
This can mean including a traditional random forest alongside a deep learning approach, or having multiple random forests with trees of different depths.
Throughout this process, the explainability of various models should be kept in mind.
Some applications may be suitable for black-box predictions and value predictive performance above all else, whereas others may require a transparent decision-making process, even if it results in less accurate predictions.

\textbf{Example.}
In the PECARN ciTBI task, it is important that doctors understand why the model is recommending forgoing a CT scan.
Thus, it is essential to choose a clear and intuitive algorithm to ensure successful deployment.
As a result, the clinical decision rule (CDR) is a decision tree with only a few steps, ensuring ease-of-use for clinicians and explainability for patients.
The PECARN team then evaluates this CDR in a PCS-adherent way by reporting multiple performance metrics.
\subsection{PCS Prediction Results}
\label{subsec:pred-results}

After completing model training, we have a collection of several ``predictive fits".
These fitted algorithms are the result of different judgment calls, including distinct training sets, hyperparameter choices, and more.
The data scientist now must choose how predictions will be generated.
If the applied problem requires very clear and transparent predictions, then selecting a single model from this collection would be the most appropriate choice.
However, there have been recent calls to abandon the oft-used single model framework in favor of ensembles which may pick up on different types of signal \cite{rudin2024models}.
To form such an ensemble, a handful of predictive fits should be selected via a prediction-check on held-out data.
A suitable threshold for ``good" performance may change based on the application, but possible rules could be to select the top $k\%$ of models or to select all models with some small $\varepsilon$ accuracy of the best-performing model.
This type of ensemble is aligned with Leo Breiman's idea of the \textit{Rashomon Effect}, which describes the phenomenon where many equally-performing models exist for the same task \cite{breiman2001cultures}.

Given either a single predictive fit or a ``PCS ensemble" consisting of well-performing models, we now must generate predictions for new data points.
Obtaining a point estimate is typically straightforward.
However, a single predicted value does not display the entire picture.
We argue that when possible, some form of uncertainty quantification should be provided alongside the point estimate.
In particular, we recommend using PCS prediction perturbation intervals \cite{veridical2024yu, agarwal2025pcsuq}.
Prediction perturbation intervals represent a range of plausible predictions from models which passed the predictability-check.
For single-model use cases, this can correspond to models trained on different training sets, with possible perturbations including bootstrapped samples or different upstream judgment calls.
Prediction tasks with a continuous outcome can then take a lower and upper quantile (e.g. 0.05 and 0.95) from the predicted responses to create the interval.
For more information on prediction perturbation intervals, including calibration and coverage information, we refer the reader to \href{https://vdsbook.com/13-final_ml}{Chapter 13} of \citet{veridical2024yu} and \citet{agarwal2025pcsuq}.

\textbf{Example.}
The ciTBI task provides a good example of a case where creating an ensemble is not feasible for the applied problem.
Recall that the goal was to create a CDR to help emergency department clinicians make decisions regarding treatment.
These CDRs must be incredibly interpretable and easy-to-use, as these decisions are made quickly by the clinician on duty.
For the problem of recommending CT scans to screen for ciTBI, the researchers arrived at an inherently explainable tree, which is visualized in Figure 3 of \citet{kuppermann2009tbi}.
Ensembling this tree with other predictive fits would make it harder to apply quickly and explain to patients, making the single-model approach better adapted for this task.

\begin{center}
    \begin{tcolorbox}[colback=yellow!20, colframe=black, width=0.9\linewidth]
    \centering\faLightbulb[regular]\hspace{0.25cm}\textbf{Generative AI Tips \& Warnings}\hspace{0.25cm}\faExclamationTriangle
        \begin{itemize}[leftmargin=0pt]
            \item Do not allow GenAI to create the entire modeling pipeline at once. Instead, break it into smaller modules, such as parameter tuning and model selection.
            \begin{itemize}
                \item Always review the generated code for correctness, efficiency, and alignment with the problem context.
            \end{itemize}
            \item Tools may forget to use random seeds, harming reproducibility of predictions.
        \end{itemize}
    \end{tcolorbox}
\end{center}
\begin{center}
    \begin{tcolorbox}[width=0.9\linewidth]
    \centering\faClipboardList\hspace{0.25cm}\textbf{Key Takeaways \& Action Items}\hspace{0.25cm}\faClipboardCheck
        \begin{itemize}[leftmargin=0pt]
            \item Stability is encouraged in model development through perturbations to both data and algorithms.
            \item Rather than relying on a single ``optimal" model, the PCS framework encourages using multiple models when feasible.
            \item The well-performing models can be combined into a ``PCS ensemble" to create more stable point estimates as well as prediction perturbation intervals.
        \end{itemize}
    \end{tcolorbox}
\end{center}

\section{DSLC Stage \#5: Evaluation of Results}
\label{sec:eval}

In practice, scrutinizing your findings should be practiced throughout the DSLC.
However, due to its importance, we give it special consideration here.
When evaluating the findings from data analysis or results of predictions, it is important to be wary of confirmation bias.
Intentional measures should be taken to ensure confirmation bias is not changing how the results are interpreted.
One suggestion is to use ``dummy" labels when comparing results to competing methods or approaches.
Depending on the type of finding, it may also be useful to show ``fake" versions of results to domain experts alongside the actual results, allowing the experts to identify that the real findings make sense \cite{veridical2024yu}.
This allows collaborators, domain experts, and the researchers to determine that the actual conclusions are trustworthy and offer a noticeable improvement over competing approaches.

Another way to bring stability checks into the scrutinization step is to check that results are consistent across different types of reasonable visualizations.
For example, two distributions might seem to have different shapes when shown via a violin plot, but a boxplot may present a different angle.
Furthermore, depending on the domain problem, it may help to check that the results are consistent across different reasonable metrics.
For example, it may be beneficial to plot both the raw frequency of an outcome and its rate, rather than just one or the other.

\textbf{Example.}
The PECARN team does not compare their performance to existing baselines, likely because no universal baselines had been established.
They present two comprehensive tables of results, organized by categories of key features.
These tables include uncertainty quantification, offering valuable insight into model robustness.
In addition, their Figure 2 provides a visual representation of these results, serving as an alternative way to inspect performance.

\begin{center}
    \begin{tcolorbox}[colback=yellow!20, colframe=black, width=0.9\linewidth]
    \centering\faLightbulb[regular]\hspace{0.25cm}\textbf{Generative AI Tips \& Warnings}\hspace{0.25cm}\faExclamationTriangle
        \begin{itemize}[leftmargin=0pt]
            \item GenAI can help create dummy labels and altered plots for comparison.
            \item Coding assistants can generate different types of plots to verify consistency.
            \item Do not allow LLMs to draw conclusions from any visualizations. This is the job of the analyst and collaborating domain experts.
        \end{itemize}
    \end{tcolorbox}
\end{center}
\begin{center}
    \begin{tcolorbox}[width=0.9\linewidth]
    \centering\faClipboardList\hspace{0.25cm}\textbf{Key Takeaways \& Action Items}\hspace{0.25cm}\faClipboardCheck
        \begin{itemize}[leftmargin=0pt]
            \item Compare results to competing methods and approaches without knowing which is which, thus helping avoid confirmation bias.
            \item Ensure that results are stable across reasonable visualization schemes.
        \end{itemize}
    \end{tcolorbox}
\end{center}

\section{DSLC Stage \#6: Communication of Results}
\label{sec:communication}

The final step in the DSLC is sharing your findings with your target audience.
Depending on the context, this could mean a presentation to company leadership, creating software, or publishing a paper.
The resulting product should be tailored to create a frictionless way for your work to be used in making real-world decisions.
Thus, be careful assuming others have prior knowledge about your project.
While the takeaways or conclusions from a certain result may seem obvious, it is still recommended to be explicit about its interpretation and results.

The true goal for most data science projects is for the results to be used for real-world decision making, which we call being put ``into production".
Even for academic projects resulting in publications, this should mean making any code publicly available and easy-to-use.
For example, a novel predictive algorithm for detecting melanoma will not have much real-world impact if it is only described in a paper.
Clinicians will not have the technical training or time to download messy code and get it working with their patients' data.
It would be more practical to incorporate the algorithm into existing clinical software or make a web interface for easy access.

The extra effort necessary to make this happen can be a lot of work for data scientists, especially considering many do not have the skills or experienced needed to make easy-to-use software.
We recommend data scientists in large organizations or companies collaborate with staff software engineers to create better tools for the adoption of results and findings.
For others without such resources on hand, an attainable goal is to create a well-documented GitHub repository with all relevant code, perhaps paired with an interactive web application using tools such as Shiny \cite{chang2024shiny}.

\textbf{Example.}
In the case of the PECARN ciTBI clinical decision rule, the authors published their findings in a peer-reviewed journal.
Furthermore, they made their dataset publicly available, including thorough data documentation as well as the forms used for entry.
However, the PECARN team did not publish any code for data cleaning or modeling, making it difficult to evaluate judgment calls made in these stages.

\begin{center}
    \begin{tcolorbox}[colback=yellow!20, colframe=black, width=0.9\linewidth]
    \centering\faLightbulb[regular]\hspace{0.25cm}\textbf{Generative AI Tips \& Warnings}\hspace{0.25cm}\faExclamationTriangle
        \begin{itemize}[leftmargin=0pt]
            \item Language models can draft documentation for code and/or data repositories.
            \item Individuals with limited programming experience can use coding assistants to create simple web applications with demonstrations of their conclusions.
            \item As always, GenAI produced documentation must be checked and verified.
        \end{itemize}
    \end{tcolorbox}
\end{center}
\begin{center}
    \begin{tcolorbox}[width=0.9\linewidth]
    \centering\faClipboardList\hspace{0.25cm}\textbf{Key Takeaways \& Action Items}\hspace{0.25cm}\faClipboardCheck
        \begin{itemize}[leftmargin=0pt]
            \item Findings should be presented in a way that makes them easy to understand and apply---whether it's through presentations, software, or publications.
        \end{itemize}
    \end{tcolorbox}
\end{center}

\section{Case Study: Downstream Effects of Data Cleaning}\label{sec:casestudy}

Uncertainty stemming from different data cleaning processes is underemphasized in data science.
To demonstrate how choices in data cleaning can change results in other DSLC stages, we explore a real-world case study \footnote{\fontsize{7.25pt}{10pt}\selectfont Inspired by Omer Ronen, who performed a similar study as the TA for the PhD applied statistics course at UC Berkeley.}\textsuperscript{,}\footnote{\fontsize{7.25pt}{10pt}\selectfont All code and data can be found at \faGithub\ \url{https://github.com/Yu-Group/pcs-workflow}.}.
During the Spring 2025 semester, the second author taught a graduate statistics course (STAT 214) at UC Berkeley called ``Data Analysis and Machine Learning for Real-World Decision Making".
The first author served as a TA with two others.
The main task of the first class project was to thoroughly explore and clean the PECARN ciTBI dataset.
All students received the same documentation and instructions, including advice from Dr. Aaron Kornblith, a pediatric emergency medicine doctor affiliated with PECARN.
Using their cleaned data, the students were then asked to model a potential decision rule for recommending CT scans.
Similar to the goal of \citet{kuppermann2009tbi}, these recommendations should continue providing scans for children at risk of ciTBI while minimizing the total number of CT scans given.
Of the 57 students enrolled in this course, 19 gave us explicit permission to use their cleaned datasets in this work.

\subsection{Data Description}

The raw ciTBI dataset consists of 125 covariates measured across over 43,000 patients.
These covariates can be divided into four types: logistical information, important symptoms, treatment received, and patient outcomes.
We drop the logistical covariates such as patient indentifiers, as they are not relevant to the prediction task.
Similarly, we ignore information regarding administered treatment, as this provides proxy outcomes which interfere with our task of recommending CT scans.
We focus particularly on the outcome of ciTBI diagnosis, attempting to use information on relevant symptoms to uncover data-driven insights.
We note that to maximize the number of ciTBIs detected via CT scans, we need to minimize the false negative rate (FNR).

\subsection{Judgment Calls}
\label{subsec:judgment-calls}

\begin{minipage}{0.44\textwidth}
Even with instructions, the students had to make numerous judgment calls while cleaning the data.
Perhaps the most impactful judgment calls were the handling of incomplete or invalid entries.
As previously mentioned in Section \ref{sec:cleaning}, both intentional and unintentional missingness are present in the ciTBI data.
Similiarly, many rows also contained invalid or contradictory entries.
For example, a commonly used diagnostic score for head trauma is the Glasgow Coma Score (GCS), which is defined as the sum of three subscores.
Some patients are recorded as having GCS scores which are not equivalent to the total of their subscores, denoting a possible data entry error.
Choosing whether to drop or how to impute impacted observations is a crucial judgment call for downstream modeling.
\end{minipage}
\hfill
\begin{minipage}{0.54\textwidth}
  \centering
  \includegraphics[width=\linewidth]{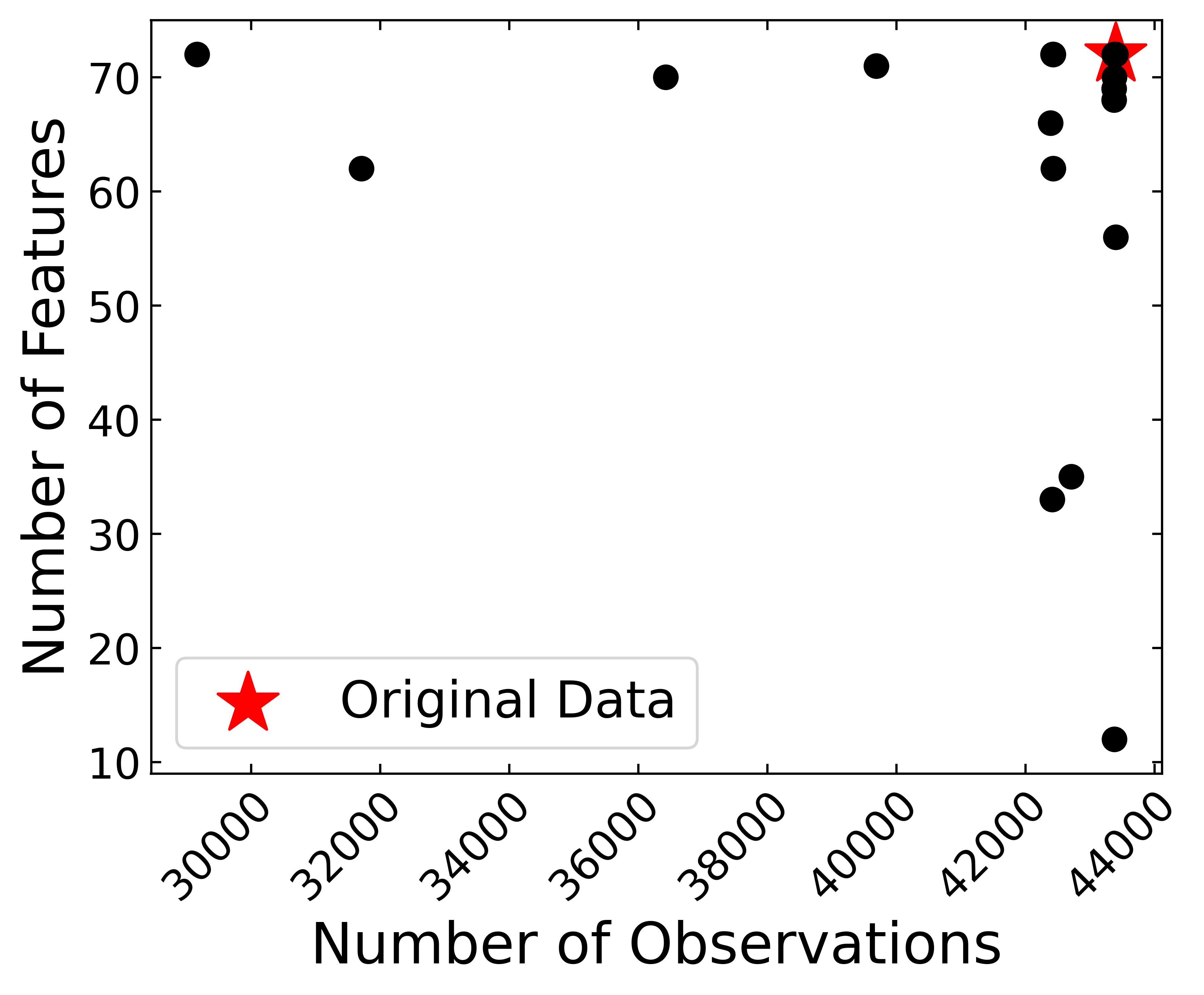}
  \captionof{figure}{Number of features and observations in the students' cleaned datasets. The students took a wide variety of approaches, with some dropping nearly a third of the observations and others selecting a small subset of features.}
  \label{fig:data-size}
\end{minipage}

We observe in Figure \ref{fig:data-size} that the students took wildly different approaches when handling these cases, with some students selecting less than 20\% of the features to retain and others dropping nearly a third of the patients from the dataset.
Therefore, our primary focus will be the treatment of missing and invalid data, encapsulating dropping or imputing observations as well as selecting particular covariates.
While ensembling these heterogeneously processed datasets would provide an avenue for improved predictions, the varying approaches to missingness creates challenges for combined inference and validation.
\href{https://vdsbook.com/13-final_ml}{Chapter 13} of \citet{veridical2024yu} details the construction of PCS ensembles and prediction perturbation intervals for differing workflows, but more work is needed to properly combine estimators fit on similar but distinct datasets.

\subsubsection{The Treatment of Missing Values Contributes to Predictive Instability}
\label{subsubsec:missing-vals}

To evaluate the ramifications of data dropping and imputation, we implement the clinical decision rule (CDR) proposed by \citet{kuppermann2009tbi}.
For children under two years of age, this CDR recommends a CT scan if there are signs of altered mental status or basiliar skull fracture.
For children two and older, CT is recommended when there are signs of altered mental status or palpable skull fracture.
This decision rule is selected to evaluate results because it uses a small set of variables which all 19 students kept.

We use the CDR to obtain CT recommendations for all 19 cleaned datasets.
This produced a wide range of false negative rates, with the lowest being 0.05\% and the highest being 0.38\%.
Then, for $i=1,\dots,19$, we computed 100 bootstrapped versions of student $i$'s cleaned data.
CT recommendations for the bootstrapped samples were generated using the same CDR.
The resulting distributions of false negative rates are shown in Figure \ref{fig:cdr-uncertainty}.

\noindent
\begin{minipage}{0.35\textwidth}
We observe that the uncertainty resulting from data cleaning judgment calls is on the same order of magnitude as the uncertainty induced by bootstrap sampling.
In fact, the distribution of false negative rates resulting from different judgment calls has greater variance than that of any of the bootstrapped datasets, in part due to the presence of large outliers.
This provides strong evidence that ignoring variability from judgment calls may result in underestimating uncertainty, harming the trustworthiness of conclusions.
\end{minipage}
\hfill
\begin{minipage}{0.6\textwidth}
  \centering
  \includegraphics[width=\linewidth]{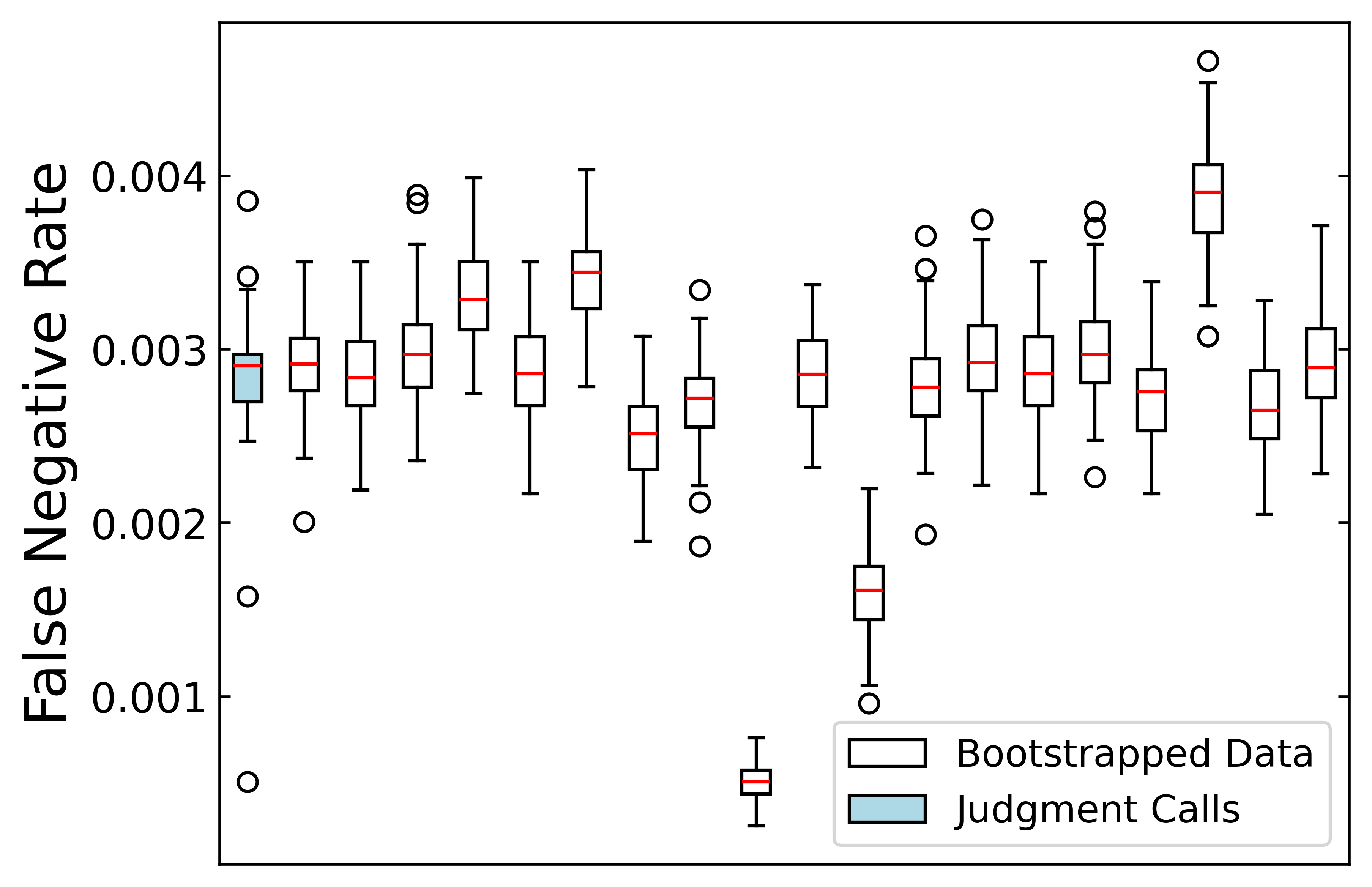}
  \captionof{figure}{Uncertainty in CDR predictions resulting from judgment calls. Note that the variability of the false negative rate stemming from judgment calls is comparable to that of the bootstrapped datasets.}
    \label{fig:cdr-uncertainty}
\end{minipage}

\subsubsection{The Selection of Covariates for Modeling Increases Predictive Uncertainty}

Figure \ref{fig:data-size} shows that students also made a wide variety of choices when deciding which variables may be helpful for downstream modeling.
Since the CDR only uses a small number of variables which each cleaned dataset has in common, it does not capture the uncertainty resulting from the inclusion of different features.
To investigate this uncertainty, we fit a logistic regression model to each student's cleaned dataset.
Specifically, we use a 70/30 train-test split, fitting the regression model to the training set and evaluating performance on the held-out test set.
We set the threshold for CT scan recommendation at 1/313.
This number is selected since it represents the lifetime attributable risk of cancer due to head CT scans on children \cite{smith-bindman2025cancer}.
Therefore, this threshold corresponds to recommending CT if the probability of ciTBI is greater than the lifetime attributable risk of cancer, and avoiding CT otherwise.
We note that because logistic regression cannot be performed in the presence of missingness, four cleaned datasets had to be further preprocessed.
For these datasets, we first drop all columns with more than 10,000 NA values.
These dropped features largely represent descriptors for other covariates.
We then drop any remaining patients with incomplete entries.

CT recommendations using the logistic regression model were first obtained for all 19 cleaned datasets.
We then followed the bootstrap procedure described above in Section \ref{subsubsec:missing-vals}, fitting a logistic regression model on the training data for each bootstrap sample and generating predictions on the held-out samples.
The resulting distributions of false negative rates can be seen in Figure \ref{fig:log-uncertainty}.

\noindent
\begin{minipage}{0.37\textwidth}
We observe a wide array of variation across the bootstrapped datasets, with a large amount of uncertainty stemming from judgment calls.
We note that IQR of FNRs resulting from different judgment calls is greater than that of 14 out of 19 bootstrapped datasets, indicating that decisions such as selecting different features adds substantative uncertainty into the downstream predictions.
This provides evidence that frameworks such as PCS are needed to quantify and address the uncertainty arising from judgment calls in the DSLC.
\end{minipage}
\hfill
\begin{minipage}{0.59\textwidth}
  \centering
  \includegraphics[width=\linewidth]{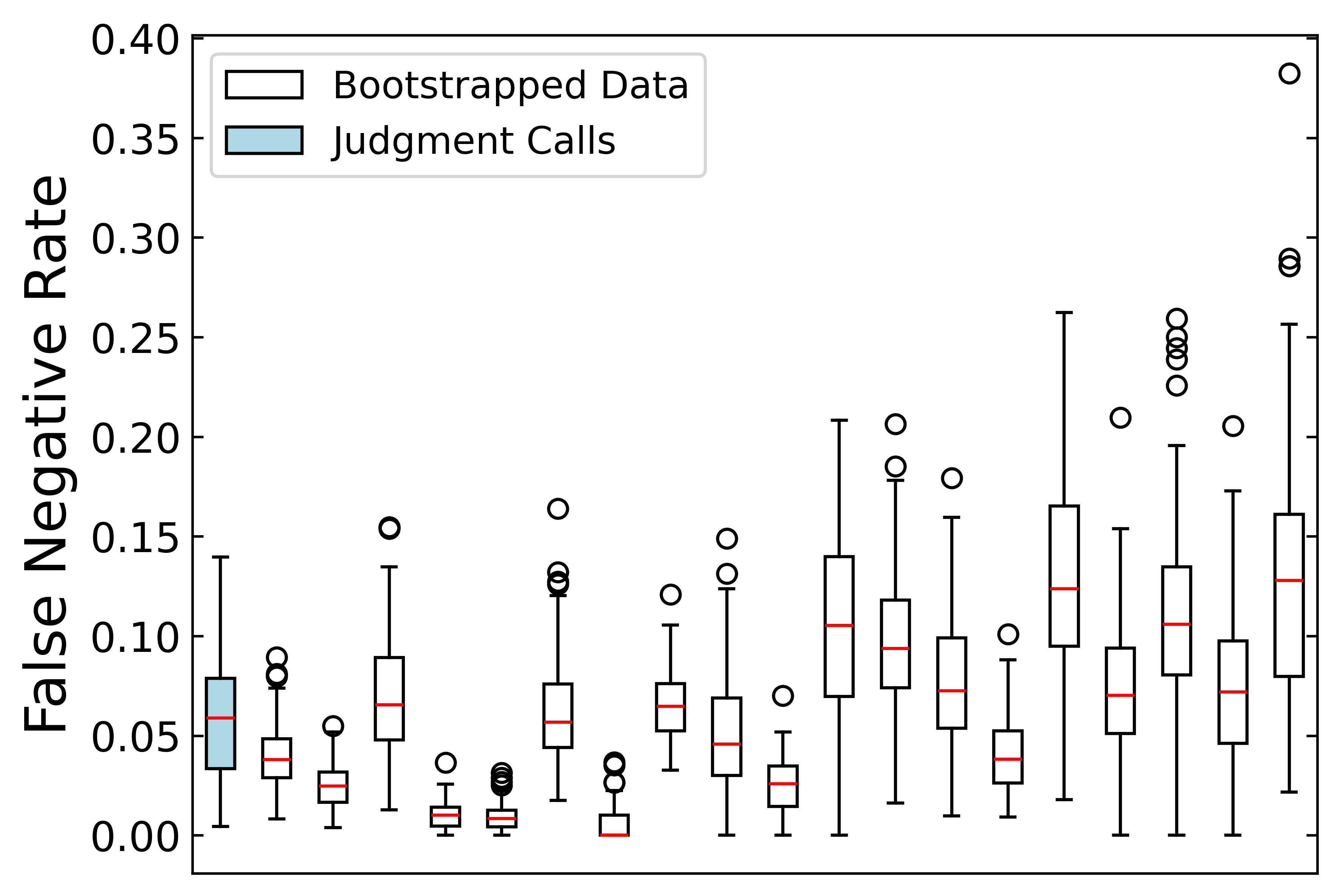}
  \captionof{figure}{Logistic regressions fit on each of the 19 cleaned datasets vary widely with respect to the resulting FNRs.}
    \label{fig:log-uncertainty}
\end{minipage}

\section{Discussion}\label{sec:discussion}

The PCS workflow for veridical data science (VDS) provides a practical approach to create replicable results through systematically addressing uncertainty caused by judgment calls in the DSLC.
By incorporating tools and practices that check reality and quantify both stability and uncertainty in a computationally efficient manner, PCS enables more reliable and truthful data science.
The PCS principles are broadly applicable across domains where data science and AI are used to draw conclusions, such as the social sciences and biomedical sciences.
Although this paper focuses on synthesizing the current practice of PCS while adding GenAI guidance, the framework is still evolving and has many possible extensions.
Four such extensions are discussed briefly below.

\textbf{(1) Mitigating Difficulties of Implementing the PCS Workflow.}
While we maintain that the PCS principles are universally applicable, it is important to acknowledge that implementing the current workflow can require a non-trivial amount human effort and computational resources.
Although some implementations can be rather straightforward, such as obtaining PCS rankings of gene expression features \cite{wang2024epistasis}, keeping track of results under different judgment calls grows exponentially \cite{gelman2013garden}.
For this reason, adhering to the entire PCS workflow may not be feasible in certain settings.
Specialized software which tracks and manages judgment calls would make the implementation of PCS much easier.
The Yu Group has begun addressing this bottleneck \cite{duncan2022veridicalflow, Duncan2024simChef, elliott2024designing}.
We are optimistic about a wider adoption of PCS in the near future while recognizing that more work is needed to fully overcome these challenges.

\textbf{(2) PCS GenAI Benchmarking.}
As GenAI capabilities continue to advance, companies and researchers are beginning to develop data science `agents' that can automate complex data analysis tasks and even generate actionable conclusions \cite{sun2025agenticdataagenticdataanalytics, fine2025datascienceagent}.
This has resulted in a growing body of benchmarking research which aims to evaluate the performance of these agents on real-world data science workflows \cite{gu2024blade, majumderdiscoverybench}.
We believe that the PCS workflow naturally lends itself to such a purpose, as agents can be evaluated on the predictability and stability of their findings under different judgment calls as well as the produced code's adherence to the MERITS criteria \cite{elliott2024designing}.
For example, benchmarking could include applying agents to datasets with predefined perturbations or alternate modeling choices, with each agent being scored on the consistency of their conclusions across variations.
Furthermore, much effort has been put into developing GenAI tools with `mathematical reasoning' capabilities, but very little effort has been given to statistical reasoning.
We see a PCS benchmarking process as having the potential to ensure trustworthy agent findings, as well as a way to evaluate the statistical reasoning abilities of GenAI systems.

\textbf{(3) PCS Design.}
One active research direction involves creating PCS-adherent strategies for generalized experimental design \cite{murphy2024modernizing}.
Whereas traditional experimental design often relies on expensive and time-consuming controlled experiments to establish causality, PCS design emphasizes carefully structuring the DSLC to arrive at cost-effective decisions.
For instance, \citet{wang2024epistasis} is able to identify genetic drivers of cardiac hypertrophy by providing interpretable hypothesis generation recommendations for further experiments.
Further works may include treatment effect estimation using prediction-checked adjustments or extending PCS to Bayesian design settings.

\textbf{(4) PCS Methodology.}
The PCS framework encourages practitioners to select reality and stability-checked models.
While certain out-of-the-box models like random forests may sometimes provide predictive stability, certain applications may require developing new inherently stable and interpretable methodology.
Examples include \citet{dwivedi2020stable}, which introduces a method for discovering stable subgroups in randomized experiments, and \citet{basu2018iterative}, which proposes an algorithm for identifying predictive and stable high-order interactions.
Any applied problem with a large amount of uncertainty from judgment calls may benefit from similar PCS-aligned modeling algorithms.

We hope that this paper will encourage practitioners and researchers to try the PCS framework for their problems of interest.
For those who would like to learn more, we recommend watching the recordings of the 2024 \href{https://na.eventscloud.com/website/69057/home/}{Berkeley-Stanford Workshop on VDS} or gathering information on other workshops such as the June 2025 \href{https://www.integreat.no/events/public-events/workshops/veridical-data-science.html}{Rome Workshop on VDS} and July 2025 \href{https://www.eventbrite.com/e/veridical-data-science-for-biology-2025-tickets-1384456339179}{VDS Workshop for Biology} by QB3 at UC Berkeley.
Materials (including code) for incorporating PCS principles into data-related project can be found on the Yu Group \href{https://www.stat.berkeley.edu/~yugroup/}{website} and \href{https://github.com/Yu-Group}{GitHub}. New PCS developments, including a biweekly talk series on Veridical Data Science in Biology (in the planning phase), will be available at the second author's \href{https://binyu.stat.berkeley.edu}{website}.

\paragraph{Acknowledgments.} The authors thank the following UC Berkeley STAT 214 (Spring 2025) students for granting us permission to use their cleaned datasets: Brian Fernando, Connor Pestell, Dominic Fannjiang, Tristan Erz, Xiaoyi Chen, Jaeyeon Lee, Joergen Jore, Junya Tsuneishi, Celestie Okechukwu, Luna Kim, Zhaoyi Zhang, Xinqi Qian, Ruiwen Liu, Tianrui Yang, Adhith Pisipati, Yuyang Wu, Ashley Zhang, Yirong Huo, and Yuki Asahara. The authors would also like to thank Anthony Ozerov, Nicolas Sanchez, and Anqi Wang for their helpful comments. ZTR is supported by the National Science Foundation Graduate Research Fellowship Program under Grant No. DGE-2146752. BY gratefully acknowledges partial support from NSF grant DMS-2413265, NSF grant DMS 2209975, NSF grant 2023505 on Collaborative Research: Foundations of Data Science Institute (FODSI), the NSF and the Simons Foundation for the Collaboration on the Theoretical Foundations of Deep Learning through awards DMS-2031883 and 814639, NSF grant MC2378 to the Institute for Artificial CyberThreat Intelligence and OperatioN (ACTION), and NIH (DMS/NIGMS) grant R01GM152718. Any opinions, findings, and conclusions or recommendations expressed in this material are those of the authors and do not necessarily reflect the views of the funding agencies.

\clearpage
\bibliographystyle{unsrtnat}
\bibliography{references}

@article{uncertainty2023yu,
title = {What is uncertainty in today’s practice of data science?},
journal = {Journal of Econometrics},
volume = {237},
number = {1},
pages = {105519},
year = {2023},
issn = {0304-4076},
doi = {https://doi.org/10.1016/j.jeconom.2023.105519},
url = {https://www.sciencedirect.com/science/article/pii/S030440762300235X},
author = {Yu, Bin}
}

@article{
vds2020yu,
author = {Bin Yu  and Karl Kumbier},
title = {Veridical data science},
journal = {Proceedings of the National Academy of Sciences},
volume = {117},
number = {8},
pages = {3920-3929},
year = {2020},
doi = {10.1073/pnas.1901326117},
URL = {https://www.pnas.org/doi/abs/10.1073/pnas.1901326117},
eprint = {https://www.pnas.org/doi/pdf/10.1073/pnas.1901326117}}

@article{stability2013yu,
author = {Bin Yu},
doi = {10.3150/13-BEJSP14},
journal = {Bernoulli},
number = {4},
pages = {1484 -- 1500},
publisher = {Bernoulli Society for Mathematical Statistics and Probability},
title = {{Stability}},
url = {https://doi.org/10.3150/13-BEJSP14},
volume = {19},
year = {2013}}

@book{veridical2024yu,
author = {Yu, Bin and Barter, Rebecca L.},
title = {Veridical Data Science: The Practice of Responsible Data Analysis and Decision Making},
publisher = {MIT Press},
year = {2024},
url = {https://vdsbook.com/}}

@inproceedings{
chiang2024chatbot,
title={Chatbot Arena: An Open Platform for Evaluating {LLM}s by Human Preference},
author={Wei-Lin Chiang and Lianmin Zheng and Ying Sheng and Anastasios Nikolas Angelopoulos and Tianle Li and Dacheng Li and Banghua Zhu and Hao Zhang and Michael Jordan and Joseph E. Gonzalez and Ion Stoica},
booktitle={Forty-first International Conference on Machine Learning},
year={2024},
url={https://openreview.net/forum?id=3MW8GKNyzI}
}

@Manual{chang2024shiny,
    title = {shiny: Web Application Framework for R},
    author = {Winston Chang and Joe Cheng and JJ Allaire and Carson Sievert and Barret Schloerke and Yihui Xie and Jeff Allen and Jonathan McPherson and Alan Dipert and Barbara Borges},
    year = {2024},
    note = {R package version 1.8.1.1},
    url = {https://CRAN.R-project.org/package=shiny}
  }

@inproceedings{rudin2024models,
author = {Rudin, Cynthia and Zhong, Chudi and Semenova, Lesia and Seltzer, Margo and Parr, Ronald and Liu, Jiachang and Katta, Srikar and Donnelly, Jon and Chen, Harry and Boner, Zachery},
title = {Position: amazing things come from having many good models},
year = {2024},
publisher = {JMLR.org},
booktitle = {Proceedings of the 41st International Conference on Machine Learning},
articleno = {1742},
numpages = {13},
location = {Vienna, Austria},
series = {ICML'24}
}

@article{breiman2001cultures,
 author = {Leo Breiman},
 journal = {Statistical Science},
 number = {3},
 pages = {199--215},
 publisher = {Institute of Mathematical Statistics},
 title = {Statistical Modeling: The Two Cultures},
 urldate = {2025-04-22},
 volume = {16},
 year = {2001}
}

@article{kuppermann2009tbi,
author={Kuppermann, Nathan
and Holmes, James F.
and Dayan, Peter S.
and Hoyle, John D.
and Atabaki, Shireen M.
and Holubkov, Richard
and Nadel, Frances M.
and Monroe, David
and Stanley, Rachel M.
and Borgialli, Dominic A.
and Badawy, Mohamed K.
and Schunk, Jeff E.
and Quayle, Kimberly S.
and Mahajan, Prashant
and Lichenstein, Richard
and Lillis, Kathleen A.
and Tunik, Michael G.
and Jacobs, Elizabeth S.
and Callahan, James M.
and Gorelick, Marc H.
and Glass, Todd F.
and Lee, Lois K.
and Bachman, Michael C.
and Cooper, Arthur
and Powell, Elizabeth C.
and Gerardi, Michael J.
and Melville, Kraig A.
and Muizelaar, J. Paul
and Wisner, David H.
and Zuspan, Sally Jo
and Dean, J. Michael
and Wootton-Gorges, Sandra L.},
title={Identification of children at very low risk of clinically-important brain injuries after head trauma: a prospective cohort study},
journal={The Lancet},
year={2009},
month={Oct},
day={03},
publisher={Elsevier},
volume={374},
number={9696},
pages={1160-1170},
issn={0140-6736},
doi={10.1016/S0140-6736(09)61558-0},
url={https://doi.org/10.1016/S0140-6736(09)61558-0}
}

@article{huang2025hallucination,
author = {Huang, Lei and Yu, Weijiang and Ma, Weitao and Zhong, Weihong and Feng, Zhangyin and Wang, Haotian and Chen, Qianglong and Peng, Weihua and Feng, Xiaocheng and Qin, Bing and Liu, Ting},
title = {A Survey on Hallucination in Large Language Models: Principles, Taxonomy, Challenges, and Open Questions},
year = {2025},
issue_date = {March 2025},
publisher = {Association for Computing Machinery},
address = {New York, NY, USA},
volume = {43},
number = {2},
issn = {1046-8188},
url = {https://doi.org/10.1145/3703155},
doi = {10.1145/3703155},
journal = {ACM Trans. Inf. Syst.},
month = jan,
articleno = {42},
numpages = {55}
}

@ARTICLE{doezema1991ct,
  title     = "Magnetic resonance imaging in minor head injury",
  author    = "Doezema, D and King, J N and Tandberg, D and Espinosa, M C and
               Orrison, W W",
  journal   = "Ann. Emerg. Med.",
  publisher = "Elsevier BV",
  volume    =  20,
  number    =  12,
  pages     = "1281--1285",
  month     =  dec,
  year      =  1991,
  language  = "en"
}

@ARTICLE{faul2010tbi,
  title     = "Traumatic Brain Injury in the United States: Emergency Department Visits, Hospitalizations and Deaths 2002-2006",
  author    = "Faul, M and Xu, L and Wald, M M and Coronado, V G",
  journal   = "National Center for Injury Prevention and Control",
  publisher = "Centers for Disease Control and Prevention",
  year      =  2010
}

@ARTICLE{langlois2006tbi,
  title     = "Traumatic Brain Injury in the United States: Emergency Department Visits, Hospitalizations, and Deaths",
  author    = "Langlois, J A and Rutland-Brown, W and Thomas, K E",
  journal   = "National Center for Injury Prevention and Control",
  publisher = "Centers for Disease Control and Prevention",
  year      =  2006
}

@Manual{tangvdocs,
    title = {vdocs: Beautiful Documentation for PCS-style Analyses},
    author = {Tiffany Tang and Ana Maria Kenney},
    year = {2023},
    note = {R package version 0.0.0.9000, commit f7a22d4ab6c5c5e450ac53c71341f2451d49905e},
    url = {https://github.com/Yu-Group/vdocs},
  }

@article{ward1963cluster,
author = {Joe H. Ward Jr. and},
title = {Hierarchical Grouping to Optimize an Objective Function},
journal = {Journal of the American Statistical Association},
volume = {58},
number = {301},
pages = {236--244},
year = {1963},
publisher = {ASA Website},
doi = {10.1080/01621459.1963.10500845},
URL={https://www.tandfonline.com/doi/abs/10.1080/01621459.1963.10500845},
eprint = {https://www.tandfonline.com/doi/pdf/10.1080/01621459.1963.10500845}
}

@article{lloyd1982least,
  title={Least squares quantization in PCM},
  author={Lloyd, Stuart},
  journal={IEEE transactions on information theory},
  volume={28},
  number={2},
  pages={129--137},
  year={1982},
  publisher={IEEE}
}

@inproceedings{macqueen1967some,
  title={Some methods for classification and analysis of multivariate observations},
  author={MacQueen, James},
  booktitle={Proceedings of the Fifth Berkeley Symposium on Mathematical Statistics and Probability, Volume 1: Statistics},
  volume={5},
  pages={281--298},
  year={1967},
  organization={University of California press}
}

@article{vandermaaten2008tsne,
  author  = {Laurens van der Maaten and Geoffrey Hinton},
  title   = {Visualizing Data using t-SNE},
  journal = {Journal of Machine Learning Research},
  year    = {2008},
  volume  = {9},
  number  = {86},
  pages   = {2579--2605},
  url     = {http://jmlr.org/papers/v9/vandermaaten08a.html}
}

@ARTICLE{BenHur2002stability,
  title    = "A stability based method for discovering structure in clustered
              data",
  author   = "Ben-Hur, Asa and Elisseeff, Andre and Guyon, Isabelle",
  journal  = "Pac. Symp. Biocomput.",
  pages    = "6--17",
  year     =  2002,
  language = "en"
}

@Article{begley2012replication,
author={Begley, C. Glenn
and Ellis, Lee M.},
title={Raise standards for preclinical cancer research},
journal={Nature},
year={2012},
month={Mar},
day={01},
volume={483},
number={7391},
pages={531-533},
issn={1476-4687},
doi={10.1038/483531a},
url={https://doi.org/10.1038/483531a}
}

@article{breznau2022data,
author = {Nate Breznau  and Eike Mark Rinke  and Alexander Wuttke  and Hung H. V. Nguyen  and Muna Adem  and Jule Adriaans  and Amalia Alvarez-Benjumea  and Henrik K. Andersen  and Daniel Auer  and Flavio Azevedo  and Oke Bahnsen  and Dave Balzer  and Gerrit Bauer  and Paul C. Bauer  and Markus Baumann  and Sharon Baute  and Verena Benoit  and Julian Bernauer  and Carl Berning  and Anna Berthold  and Felix S. Bethke  and Thomas Biegert  and Katharina Blinzler  and Johannes N. Blumenberg  and Licia Bobzien  and Andrea Bohman  and Thijs Bol  and Amie Bostic  and Zuzanna Brzozowska  and Katharina Burgdorf  and Kaspar Burger  and Kathrin B. Busch  and Juan Carlos-Castillo  and Nathan Chan  and Pablo Christmann  and Roxanne Connelly  and Christian S. Czymara  and Elena Damian  and Alejandro Ecker  and Achim Edelmann  and Maureen A. Eger  and Simon Ellerbrock  and Anna Forke  and Andrea Forster  and Chris Gaasendam  and Konstantin Gavras  and Vernon Gayle  and Theresa Gessler  and Timo Gnambs  and Amélie Godefroidt  and Max Grömping  and Martin Groß  and Stefan Gruber  and Tobias Gummer  and Andreas Hadjar  and Jan Paul Heisig  and Sebastian Hellmeier  and Stefanie Heyne  and Magdalena Hirsch  and Mikael Hjerm  and Oshrat Hochman  and Andreas Hövermann  and Sophia Hunger  and Christian Hunkler  and Nora Huth  and Zsófia S. Ignácz  and Laura Jacobs  and Jannes Jacobsen  and Bastian Jaeger  and Sebastian Jungkunz  and Nils Jungmann  and Mathias Kauff  and Manuel Kleinert  and Julia Klinger  and Jan-Philipp Kolb  and Marta Kołczyńska  and John Kuk  and Katharina Kunißen  and Dafina Kurti Sinatra  and Alexander Langenkamp  and Philipp M. Lersch  and Lea-Maria Löbel  and Philipp Lutscher  and Matthias Mader  and Joan E. Madia  and Natalia Malancu  and Luis Maldonado  and Helge Marahrens  and Nicole Martin  and Paul Martinez  and Jochen Mayerl  and Oscar J. Mayorga  and Patricia McManus  and Kyle McWagner  and Cecil Meeusen  and Daniel Meierrieks  and Jonathan Mellon  and Friedolin Merhout  and Samuel Merk  and Daniel Meyer  and Leticia Micheli  and Jonathan Mijs  and Cristóbal Moya  and Marcel Neunhoeffer  and Daniel Nüst  and Olav Nygård  and Fabian Ochsenfeld  and Gunnar Otte  and Anna O. Pechenkina  and Christopher Prosser  and Louis Raes  and Kevin Ralston  and Miguel R. Ramos  and Arne Roets  and Jonathan Rogers  and Guido Ropers  and Robin Samuel  and Gregor Sand  and Ariela Schachter  and Merlin Schaeffer  and David Schieferdecker  and Elmar Schlueter  and Regine Schmidt  and Katja M. Schmidt  and Alexander Schmidt-Catran  and Claudia Schmiedeberg  and Jürgen Schneider  and Martijn Schoonvelde  and Julia Schulte-Cloos  and Sandy Schumann  and Reinhard Schunck  and Jürgen Schupp  and Julian Seuring  and Henning Silber  and Willem Sleegers  and Nico Sonntag  and Alexander Staudt  and Nadia Steiber  and Nils Steiner  and Sebastian Sternberg  and Dieter Stiers  and Dragana Stojmenovska  and Nora Storz  and Erich Striessnig  and Anne-Kathrin Stroppe  and Janna Teltemann  and Andrey Tibajev  and Brian Tung  and Giacomo Vagni  and Jasper Van Assche  and Meta van der Linden  and Jolanda van der Noll  and Arno Van Hootegem  and Stefan Vogtenhuber  and Bogdan Voicu  and Fieke Wagemans  and Nadja Wehl  and Hannah Werner  and Brenton M. Wiernik  and Fabian Winter  and Christof Wolf  and Yuki Yamada  and Nan Zhang  and Conrad Ziller  and Stefan Zins  and Tomasz Żółtak },
title = {Observing many researchers using the same data and hypothesis reveals a hidden universe of uncertainty},
journal = {Proceedings of the National Academy of Sciences},
volume = {119},
number = {44},
pages = {e2203150119},
year = {2022},
doi = {10.1073/pnas.2203150119},
URL = {https://www.pnas.org/doi/abs/10.1073/pnas.2203150119},
eprint = {https://www.pnas.org/doi/pdf/10.1073/pnas.2203150119}}

@ARTICLE{smith-bindman2025cancer,
  title    = "Projected Lifetime Cancer Risks From Current Computed Tomography
              Imaging",
  author   = "Smith-Bindman, Rebecca and Chu, Philip W and Azman Firdaus, Hana
              and Stewart, Carly and Malekhedayat, Matthew and Alber, Susan and
              Bolch, Wesley E and Mahendra, Malini and Berrington de
              Gonz{\'a}lez, Amy and Miglioretti, Diana L",
  journal  = "JAMA Internal Medicine",
  year     =  2025
}

@article{prinz2011believe,
  title={Believe it or not: how much can we rely on published data on potential drug targets?},
  author={Prinz, Florian and Schlange, Thomas and Asadullah, Khusru},
  journal={Nature reviews Drug discovery},
  volume={10},
  number={9},
  pages={712--712},
  year={2011},
  publisher={Nature Publishing Group UK London}
}

@article{gould2025same,
  title={Same data, different analysts: variation in effect sizes due to analytical decisions in ecology and evolutionary biology},
  author={Gould, Elliot and Fraser, Hannah S and Parker, Timothy H and Nakagawa, Shinichi and Griffith, Simon C and Vesk, Peter A and Fidler, Fiona and Hamilton, Daniel G and Abbey-Lee, Robin N and Abbott, Jessica K and others},
  journal={BMC biology},
  volume={23},
  number={1},
  pages={35},
  year={2025},
  publisher={Springer}
}

@article{elliott2024designing,
  title={Designing a Data Science simulation with MERITS: A Primer},
  author={Elliott, Corrine F and Duncan, James and Tang, Tiffany M and Behr, Merle and Kumbier, Karl and Yu, Bin},
  journal={arXiv preprint arXiv:2403.08971},
  year={2024}
}

@article{duncan2022veridicalflow,
  title={VeridicalFlow: a Python package for building trustworthy data science pipelines with PCS},
  author={Duncan, James and Kapoor, Rush and Agarwal, Abhineet and Singh, Chandan and Yu, Bin},
  journal={Journal of Open Source Software},
  volume={7},
  number={69},
  pages={3895},
  year={2022}
}

@article{jaccard1901index,
author = {Jaccard, Paul},
year = {1901},
month = {01},
pages = {547-579},
title = {Etude de la distribution florale dans une portion des Alpes et du Jura},
volume = {37},
journal = {Bulletin de la Societe Vaudoise des Sciences Naturelles},
doi = {10.5169/seals-266450}
}

@misc{agarwal2025pcsuq,
      title={PCS-UQ: Uncertainty Quantification via the Predictability-Computability-Stability Framework}, 
      author={Abhineet Agarwal and Michael Xiao and Rebecca Barter and Omer Ronen and Boyu Fan and Bin Yu},
      year={2025},
      eprint={2505.08784},
      archivePrefix={arXiv},
      primaryClass={stat.ML},
      url={https://arxiv.org/abs/2505.08784}, 
}

@article{dwivedi2020stable,
  title={Stable discovery of interpretable subgroups via calibration in causal studies},
  author={Dwivedi, Raaz and Tan, Yan Shuo and Park, Briton and Wei, Mian and Horgan, Kevin and Madigan, David and Yu, Bin},
  journal={International Statistical Review},
  volume={88},
  pages={S135--S178},
  year={2020},
  publisher={Wiley Online Library}
}

@article{wang2024epistasis,
  title={Epistasis regulates genetic control of cardiac hypertrophy},
  author={Wang, Qianru and Tang, Tiffany M and Youlton, Nathan and Weldy, Chad S and Kenney, Ana M and Ronen, Omer and Hughes, J Weston and Chin, Elizabeth T and Sutton, Shirley C and Agarwal, Abhineet and others},
  journal={medRxiv},
  pages={2023--11},
  year={2024}
}

@article{tang2025simplified,
  title={A simplified MyProstateScore2. 0 for high-grade prostate cancer},
  author={Tang, Tiffany M and Zhang, Yuping and Kenney, Ana M and Xie, Cassie and Xiao, Lanbo and Siddiqui, Javed and Srivastava, Sudhir and Sanda, Martin G and Wei, John T and Feng, Ziding and others},
  journal={Cancer Biomarkers},
  volume={42},
  number={1},
  pages={18758592241308755},
  year={2025},
  publisher={SAGE Publications Sage UK: London, England}
}

@article{gelman2013garden,
  title={The garden of forking paths: Why multiple comparisons can be a problem, even when there is no “fishing expedition” or “p-hacking” and the research hypothesis was posited ahead of time},
  author={Gelman, Andrew and Loken, Eric},
  year={2013}
}

@article{hotelling1933analysis,
	author = "Harold Hotelling",
	year = "1933",
	title = "Analysis of a complex of statistical variables into principal components",
	journal = "J. Ed. Psych.",
	volume = "24",
	pages = "417--441"
}

@article{
basu2018iterative,
author = {Sumanta Basu  and Karl Kumbier  and James B. Brown  and Bin Yu },
title = {Iterative random forests to discover predictive and stable high-order interactions},
journal = {Proceedings of the National Academy of Sciences},
volume = {115},
number = {8},
pages = {1943-1948},
year = {2018},
doi = {10.1073/pnas.1711236115},
URL = {https://www.pnas.org/doi/abs/10.1073/pnas.1711236115},
eprint = {https://www.pnas.org/doi/pdf/10.1073/pnas.1711236115}}

@article{box1976science,
 ISSN = {01621459, 1537274X},
 URL = {http://www.jstor.org/stable/2286841},
 author = {George E. P. Box},
 journal = {Journal of the American Statistical Association},
 number = {356},
 pages = {791--799},
 publisher = {[American Statistical Association, Taylor & Francis, Ltd.]},
 title = {Science and Statistics},
 urldate = {2025-05-30},
 volume = {71},
 year = {1976}
}

@article{Duncan2024simChef, doi = {10.21105/joss.06156}, url = {https://doi.org/10.21105/joss.06156}, year = {2024}, publisher = {The Open Journal}, volume = {9}, number = {95}, pages = {6156}, author = {James Duncan and Tiffany Tang and Corrine F. Elliott and Philippe Boileau and Bin Yu}, title = {simChef: High-quality data science simulations in R}, journal = {Journal of Open Source Software} }

@article{murphy2024modernizing,
    author = {Franklin, Joseph B. and Marra, Caroline and Abebe, Kaleab Z. and Butte, Atul J. and Cook, Deborah J. and Esserman, Laura and Fleisher, Lee A. and Grossman, Cynthia I. and Kass, Nancy E. and Krumholz, Harlan M. and Rowan, Kathy and Abernethy, Amy P. and JAMA Summit on Clinical Trials Participants},
    title = {Modernizing the Data Infrastructure for Clinical Research to Meet Evolving Demands for Evidence},
    journal = {JAMA},
    volume = {332},
    number = {16},
    pages = {1378-1385},
    year = {2024},
    month = {10},
    issn = {0098-7484},
    doi = {10.1001/jama.2024.0268},
    url = {https://doi.org/10.1001/jama.2024.0268},
    eprint = {https://jamanetwork.com/journals/jama/articlepdf/2822037/jama\_franklin\_2024\_sc\_240001\_1735843182.22482.pdf},
}

@misc{sun2025agenticdataagenticdataanalytics,
      title={AgenticData: An Agentic Data Analytics System for Heterogeneous Data}, 
      author={Ji Sun and Guoliang Li and Peiyao Zhou and Yihui Ma and Jingzhe Xu and Yuan Li},
      year={2025},
      eprint={2508.05002},
      archivePrefix={arXiv},
      primaryClass={cs.DB},
      url={https://arxiv.org/abs/2508.05002}, 
}

@misc{
fine2025datascienceagent,
title={Data Science Agent in Colab: The future of data analysis with Gemini},
author={Jane Fine and Mahi Kolla and Ilai Soloducho},
year={2025},
url={https://developers.googleblog.com/en/data-science-agent-in-colab-with-gemini/}
}

@inproceedings{majumderdiscoverybench,
  title={DiscoveryBench: Towards Data-Driven Discovery with Large Language Models},
  author={Majumder, Bodhisattwa Prasad and Surana, Harshit and Agarwal, Dhruv and Mishra, Bhavana Dalvi and Meena, Abhijeetsingh and Prakhar, Aryan and Vora, Tirth and Khot, Tushar and Sabharwal, Ashish and Clark, Peter},
  booktitle={The Thirteenth International Conference on Learning Representations},
  year={2025}
}

@inproceedings{gu2024blade,
  title={BLADE: Benchmarking Language Model Agents for Data-Driven Science},
  author={Gu, Ken and Shang, Ruoxi and Jiang, Ruien and Kuang, Keying and Lin, Richard-John and Lyu, Donghe and Mao, Yue and Pan, Youran and Wu, Teng and Yu, Jiaqian and others},
  booktitle={Findings of the Association for Computational Linguistics: EMNLP 2024},
  pages={13936--13971},
  year={2024}
}

\end{document}